\documentclass{article}
\usepackage{spconf,amsmath,epsfig}
\usepackage{fancyvrb}

\begin{document}\sloppy

\def\x{{\mathbf x}}
\def\L{{\cal L}}

\title{SCALABLE DOMAIN ADAPTATION OF CONVOLUTIONAL NEURAL NETWORKS}
%
\name{Adrian Popescu, Etienne Gadeski, Herv\'e Le Borgne}
\address{CEA, LIST, Vision and Content Engineering Laboratory \\
France\\ 
Email: \{adrian.popescu,etienne.gadeski,herve.le-borgne\}@cea.fr}

\maketitle

\begin{abstract}
Convolutional neural networks (CNNs) tend to become a standard approach to solve a wide array of computer vision problems. 
Besides important theoretical and practical advances in their design, their success is built on the existence of manually labeled visual resources, such as ImageNet.
The creation of such datasets is cumbersome and here we focus on alternatives to manual labeling.
We hypothesize that new resources are of uttermost importance in domains which are not or weakly covered by ImageNet, such as tourism photographs. 
We first collect noisy Flickr images for tourist points of interest and apply automatic or weakly-supervised reranking techniques to reduce noise.
Then, we learn domain adapted models with a standard CNN architecture and compare them to a generic model obtained from ImageNet.
Experimental validation is conducted with publicly available datasets, including Oxford5k, INRIA Holidays and Div150Cred. 
Results show that low-cost domain adaptation improves results compared to the use of generic models but also compared to strong non-CNN baselines such as triangulation embedding.  
\end{abstract}
\section{Introduction}
\label{sec:intro}
Many computer vision tasks, including object classification and localization, as well as content based retrieval, are increasingly tackled with convolutional neural network (CNN) architectures. 
A lot of recent papers~\cite{Chatfield:2014,sermanet:2013} focus on improving CNN architectures and assume that large-scale manually labeled datasets, such as ImageNet~\cite{deng:2009}, are readily available. However, while large, the coverage of ImageNet is insufficient in many domains. For instance, it illustrates only a very limited amount of named entities, including tourist points of interest or car brands and models. 
Such entities can nevertheless be of interest in domain-related applications and good quality visual resources which illustrate them are assumed to be necessary.
Manual resource enrichment is tedious and not considered as scalable in terms of number of concepts and images. To overcome this issue, this paper focuses on domain adaptation and more precisely on automatic or weakly-supervised methods to create the visual resources necessary for domain specific CNN training. 

Due to resource scarcity, domain transfer recently received a particular attention from the computer vision community. 
The authors of~\cite{Babenko2014} have shown that activations of top layers of a CNN can be used effectively for image retrieval. 
More interestingly, they discovered that performance increases when the network is retrained with images similar to a specific domain (tourist points of interest). 
However, their method requires important manual intervention, since they verify 200 images for each tourist point of interest (POI) included in the training set. 
Equally important, they assume that the presence of noisy images reduces the quality of CNN models and create an unbalanced training set (i.e. some POIs are illustrated by 1000 images, while a majority of them are represented by only 100 images).
In this paper, we focus on methods to build a CNN training set in a fully automated manner and show that, contrary to the hypothesis of~\cite{Babenko2014}, a CNN can actually be trained from scratch with these images. This is probably due to the intrinsic quality of the training dataset constituted with our method but also to its larger size.

The automatic collection of groups of visually coherent images was already addressed in literature. The general idea consists of collecting ``noisy'' images which are then reranked according to a learned model~\cite{schroff2011harvesting} or use clustering to determine visually compact groups~\cite{tsai2011large}. Beyond the fact these works did not aim at determining groups of image for CNN learning, we propose a simpler method, based on a finer domain characterization. Hence, our approach has a better computational tractability, while preserving CNN model accuracy. 


Experiments are carried out on three publicly available datasets which include images related to the tourism domain but have different visual properties. We test with Oxford5k~\cite{Philbin07}, INRIA Holidays~\cite{jegou:2008} and Div150Cred~\cite{Ionescu:2014} and show we obtain better results than other CNN-based method and among the best reported in the literature on these sets.

\section{Related work}
\label{sec:related}
CNNs have recently shown impressive performances in different image mining tasks~\cite{DBLP:conf/nips/KrizhevskySH12,Ciresan:2012e}.
For instance, the use of CNN  ImageNet ILSVRC challenge reduced error rate to 0.15 in 2012, 0.11 in 2013 and 0.067 in 2014~\cite{ILSVRCarxiv14} from 0.26 obtained with Fisher Vectors~\cite{perronnin10fkv}, the best performing pre-CNN features.
A wide majority of CNN related work builds on the architecture proposed in~\cite{DBLP:conf/nips/KrizhevskySH12} and proposes different improvements. 
For instance, the best performing entries of ILSVRC 2014~\cite{DBLP:journals/corr/SzegedyLJSRAEVR14,DBLP:journals/corr/SimonyanZ14a} focus on model depth and give priority to different optimizations meant to reduce the computation time. 
In parallel, a lot of attention is paid to the way available data are exploited. 
Domain transfer is another active area. The authors of~\cite{Oquab:2014} exploit CNN mid-level features learned on ImageNet data and adapt two full-connected layers to successfully classify PascalVOC 2007 data.
In a cruder approach, \cite{DBLP:journals/corr/RazavianASC14} test off-the-shelf \verb|OverFeat| features~\cite{sermanet:2013} on a wide series of benchmarks. 
They show that CNN features constitute a competitive (when not state-of-the-art) approach compared to other features which are optimized for particular datasets.
Data augmentation through simple image transformation is carefully studied in~\cite{Chatfield:2014} and leads to consequent performance improvement on the PascalVOC 2007 dataset. 
The use of the spatial structure of the image is successfully introduced in~\cite{DBLP:journals/corr/RazavianASC14} as a simple multiscale feature extraction and max-pooling scheme.
A more elaborate scheme is proposed in~\cite{gong:2014}, which exploits multiscale extraction, VLAD pooling and concatenation. 
Both~\cite{DBLP:journals/corr/RazavianASC14} and~\cite{gong:2014} report important improvement on tourism related datasets, such as INRIA Holidays and/or Oxford5k.
The work closest to ours is presented in~\cite{Babenko2014}, where authors also train a network dedicated to the tourism domain. However, their approach differs from ours in several important aspects: (1) the amount of supervision tested is significantly higher; (2) no image reranking is tested to reduce the influence of noise; (3) training is done on an highly unbalanced dataset, a modeling choice which is likely to downgrade results; (4) they initialize with the weight learned on ImageNet; (5) the data sources are different (Yandex in~\cite{Babenko2014} vs. Flickr here).
One common assumption of most CNN-related works cited above is to assume the pre-existence of a manually built visual resource good enough to train CNNs. 
In most cases, this resource is a subset of ImageNet, which is used to classify images of the same concepts~\cite{ILSVRCarxiv14} or to classify images from other datasets~\cite{DBLP:journals/corr/RazavianASC14,Oquab:2014,Chatfield:2014}.
Departing from the assumption that training data are readily available, we focus on (semi)automatic visual resource creation.
This question was already studied in non-CNN context~\cite{schroff2011harvesting,tsai2011large} but, to our knowledge, only seldom in a CNN context~\cite{Babenko2014}.
This research question has important implications because, if such resources can be built with minimal manual effort, image classification could be easily applied to a large number of domains, provided that enough data are available to describe associated concepts. 

Since the early works on efficient image retrieval~\cite{Philbin07}, many refinement were proposed such as query expansion~\cite{Chum:2011}, more efficient features~\cite{Arandjelovic12} and aggregating local features efficiently~\cite{Tolias:2013}. Some of the best results were recently obtained with  triangulation embedding~\cite{Jegou:2014}, that are particularly efficient in their compressed version, which enables scalable image search at the cost of a limited loss of performance. 

\section{CNN domain adaptation approach}
\label{sec:method}
CNNs are majoritarily used for solving computer vision problems in which train and test concepts are the same~\cite{ILSVRCarxiv14}.
However, if trained with large arrays of concepts, generic feature extractors can also be used to characterize other datasets whose concepts overlap the original ones to some extent~\cite{Oquab:2014,DBLP:journals/corr/RazavianASC14}. 
Transfer efficiency is reduced whenever the gap between train and test sets is too high and, in such cases, dedicated models should be trained. 
In this section, we describe the main steps of our domain adaptation approach which consists of the following main steps: (1) deriving relevant training concept for a given domain; (2) collecting and, in the case of noisy data, reranking training images; (3) training a CNN model with adapted data. 
During the entire process, we assume that a generic CNN models is readily available for feature extraction and use the \verb|Caffe| reference model~\cite{Jia13caffe} which is a slightly modified version of the one proposed in~\cite{DBLP:conf/nips/KrizhevskySH12}.

\subsection{Domain characterization}
One central working hypothesis here is that, when performing domain transfer with CNNs, train and test sets should be reasonably well linked. 
This link can be implemented through the use of similar concepts during training and testing. 
For instance, an image of the \textit{Empire State Building} is more likely to be well described by a CNN model obtained with a significant number of other skyscrapers among the training concepts compared to a training done with ImageNet, which contains a very low number of skyscraper related concepts.
Domain adaptation is realized through the use an external knowledge base which contains concepts from the target domain. 
For instance, if the test set contains a lot of tourism photographs, as it is the case here, we extract a list of POIs from an existing gazetteer~\cite{popescu:2008} and be used as conceptual support for adapted CNN training.
This approach assumes that an appropriate textual knowledge base is readily available for the target dataset(s). 
It would also be possible to take a data driven approach, in which images from the test set are compared to a large catalogue of visual models in order to characterize the domain.
Such a visual catalogue could be obtained, for instance, from Wikipedia but this approach falls outside of the immediate scope of this work. 

\subsection{Data collection and reranking}
The availability of large visual resources is a key component in CNN training and usage~\cite{ILSVRCarxiv14}. 
As we mentioned, when dealing with large test sets, a subset ImageNet is the preferred training resource. 
While this resource ensures large conceptual coverage, it doesn't cover domains equally well. 
Tourism is one such domain and we investigate ways to characterize it with a (semi)automatically built resource. 
Since full manual annotation of domain visual resources requires consequent efforts, we focus on low-cost alternatives.
More precisely, we investigate the following strategies:

\begin{itemize}
 \item POI-CNN-N - direct used of Flickr image sets.
 \item POI-CNN-A - automatic reranking based on the correlations between the images associated to each POI. 
 \item POI-CNN-W - weakly supervised reranking in which candidate images are compared against a limited amount of manual annotations.  
\end{itemize}
POI-CNN-N and POI-CNN-A do not require any manual intervention and are most interesting from a scalability perspective.
POI-CNN-A is implemented following the conclusions of~\cite{schroff2011harvesting}, where the authors show that reliable models can be built from limited and noisy sets of positive images. In practice, positive examples were chosen from an initial random sample of 60 images of each POI.
POI-CNN-W represents a compromise between no manual annotation and full dataset annotation, as done during the creation of ImageNet~\cite{deng:2009}.
POI-CNN-A and POI-CNN-W strategies are implemented using a linear SVM based reranking method~\cite{schroff2011harvesting}, applied to a generic CNN description of image content. 
The choice of the reranking method is done as a compromise between performance and scalability. 
To obtain a good separation of positive and negative samples, we use a 1:100 ratio between positive and negative samples. A unique negative class is defined from  500,000 diversified images. 
We use the $\ell^2$-regularized $\ell^2$-loss SVM implemented in liblinear, which is solved in the primal using a trust region Newton method. 
We determined the penalty parameter on an independent dataset (Pascal VOC) and fixed it to $C=1$ for all experiments.
For POI-CNN-A, SVM reranking is performed by using a 10-fold split of the initial positive set. 

\subsection{CNN training}
We train CNN models using the different reranking strategies presented in the preceding subsection in order to assess their influence on the quality of results. 
POI-CNN-N, POI-CNN-A and POI-CNN-W are trained with raw Flickr images, automatically reranked images and weakly supervised reranked images respectively.
To ensure comparability, the same set of POIs is used in all three cases, with the top ranked 1000 images kept for each POI. 
As we mentioned, CNN architecture optimization falls outside of our immediate scope here. 
Throughout this paper, we exploit the network architecture and training procedure proposed in~\cite{Jia13caffe} to obtain the \verb|Caffe| reference models.

\subsection{Feature tuning}
Previous works~\cite{Philbin07,Chum:2011,DBLP:journals/corr/RazavianASC14} have shown that visual instance retrieval benefits from different feature tuning methods.
However, many of these improvements, including geometric verification or recursive query expansion~\cite{Chum:2011}, imply high computational complexity and make them difficulty usable for large scale retrieval.
Since our main objective is to propose efficient and scalable features for the tourism domain, we adopt only tuning methods which keep feature computation and retrieval manageable:
\begin{itemize}
\item spatial search (S) - simple procedure described in~\cite{DBLP:journals/corr/RazavianASC14} in order to cope with object location and scale variability. Overlapping patches are extracted at different scales, with the size of each patch being $\frac{2}{l+1}\times$(width, height), where $l$ is the patch level. At each scale we extract $l^2$ patches. At even levels we also extract a patch at the center. The similarity between two patches is computed by retaining the minimum $\ell^2$ distance of the query patch to all patches of reference images. The similarity between a query and a reference image is the average of minimum distances between patches. Similar to \cite{DBLP:journals/corr/RazavianASC14}, we test different scales for query and reference images. This step has important effect on retrieval complexity since it involves $h_q \times h_r$ comparison of query and reference image patches ($h_q$ is the number of levels for query images and $h_r$ for reference images). More advanced object localization techniques exist~\cite{DBLP:journals/corr/SimonyanZ14a, sermanet:2013} but their computational cost is too high in our setting. 
\item feature augmentation (A) - pipeline including $\ell^2$ normalization, PCA-base compression, feature whitening and $\ell^2$ re-normalization introduced in \cite{Jegou:2012} and exploited in~\cite{DBLP:journals/corr/RazavianASC14}.
\item query expansion (QE) - both average query expansion method~\cite{Chum:2011} and discriminative query expansion~\cite{Arandjelovic12} were initially tested and, since we didn't find any significant difference, the first one was retained due to its lower complexity. 
This method exploit pseudo-relevance feedback to update query representation based on an initial list of top results. 
Since spatial verification is not straightforward with CNN features, it is not exploited here.
\end{itemize}

\section{Evaluation}
\label{sec:expe}
We evaluate different aspects of our approach using challenging and publicly available datasets related to the tourism domain:
\begin{itemize}
 \item \textit{Oxford5k} (Ox5k)~\cite{Philbin07} - includes photos for 11 different buildings from the Oxford area, with the buildings occupying a significant part of the photo for relevant items. All buildings have a stable and distinctive appearance, making them suitable for geometric confirmation of descriptions. 
 \item \textit{Holidays} (Holi)~\cite{jegou:2008} - is a good representative of a holiday photos folder since it includes both landscapes and POIs. Photographs are more diversified compared to \textit{Oxford5k} but the dataset is smaller.	
 \item \textit{Div150Cred}~\cite{Ionescu:2014} was very recently introduced in the context of the MediaEval 2014 evaluation exercise. The dataset includes high diversity of POI types, with variable visual complexity, is represented. \textit{Div150Cred} includes more complex objects (\textit{i.e.}\ museums, squares, parks) than \textit{Oxford5k}. Consequently, relevant images of these objects are much more diversified compared to those of buildings and the dataset is more challenging.
\end{itemize}
The usual evaluation protocols are followed for \textit{Oxford5k} and for \textit{Holidays}. 
Our approach is compared to a selection of competitive methods proposed in literature, with particular focus on mid-sized descriptors.
\textit{Div150Cred} was used to assess reranking methods for intra-POI retrieval.
We repurpose this dataset for inter-POI retrieval and use the 665 Wikipedia images provided as examples along with the collection as queries. 
Naturally, a result is considered relevant if it belongs to the same POI as a query image.
To compare our approach with existing work on this dataset, we compute results with bags of visual words with a very large vocabulary, a competitive non-CNN method and also with the \verb|Caffe| reference model, an off-the-shelf CNN feature extractor. 
Similar to the other datasets, performance is reported using Mean Average Precision (mAP)\footnote{A complete evaluation kit will be provided in order to facilitate reproducibility.}. 

All adapted CNN models are learned using the network configuration proposed in~\cite{Jia13caffe}\footnote{The lists of POIs and of images used for learning and the network configuration files will be provided in order to facilitate reproducibility.}. 
The learning process is stopped after 350,000 iterations since no significant improvement is obtained during the previous 50,000 iterations.
All results are reported with features extracted from the last fully connected layer (\verb|fc7|).
Learning was done using GTX Titan Black GPUs and took approximately one week for each model. 
Spatial search is performed with $h_q=3$ and $h_r=4$ for Oxford5k and with $h_q=2$ and $h_r=2$ for Div150Cred. 

\SaveVerb{Caffe}|Caffe|
\SaveVerb{OverFeat}|OverFeat|
\subsection{Overall results}
\begin{table}[tb!]
\centering
\caption{mAP results obtained on \textit{Oxford5k}, \textit{Holidays} and \textit{Div150Cred} datasets. T-Emb is the best configuration obtained with triangulation embedding in~\cite{Jegou:2014}. \textit{Over} and \textit{Caffe} results are obtained \protect\UseVerb{OverFeat}~\cite{sermanet:2013} and \protect\UseVerb{Caffe}~\cite{Jia13caffe} with the pretrained ImageNet CNN models provided along with these tools. CNN$_{\textrm{d}}$ is the domain adapted CNN representation introduced in~\cite{Babenko2014}. For each existing work, we report only the best performance obtained for each dataset. S, A, QE stand for spatial search, feature augmentation and query expansion respectively. POI-CNN-N, POI-CNN-A, POI-CNN-W are our adapted CNN models with no reranking, with automatic and with weakly supervised reranking. When used with subscripts, the features of our three models are compressed with PCA and the number indicates the dimension of the resulting vector. For spatial search features dimension varies because it is related to the number of patches we consider for one image. For inverted index approaches, the dimension is that of the codebook.}
\label{tab:results}
\resizebox{0.48\textwidth}{!}{
\begin{tabular}{|c|c|ccc|c|c|c|} \hline
Method & Dim. & S & A & QE & Ox5k & Holi & DIV   \\ \hline
\multicolumn{8}{|c|}{Inverted index approaches} \\ \hline
\cite{Arandjelovic12} & 1M & & & x & \textbf{92.9} & N/A & N/A  \\ \hline
\cite{Tolias:2013} & 65k & & & & 83.8 & \textbf{88} & N/A \\ \hline 
T-Emb \cite{Jegou:2014} & 8k & & & & 67.6 & 77.1 & N/A \\ \hline
BoVW & 1M & & & & 72.9 & 51.2 & 2.7 \\ \hline
 \multicolumn{8}{|c|}{Existing CNN methods} \\ \hline
\textit{Over} \cite{DBLP:journals/corr/RazavianASC14} & 4k & & & & 32.2 & 64.4 & N/A \\ \hline
\textit{Over} \cite{DBLP:journals/corr/RazavianASC14} & 4k &x &x & & 68 & 84.3 & N/A \\ \hline
MOP-CNN \cite{gong:2014} & 12k & & &  & N/A & 78.8 & N/A  \\ \hline
CNN$_{\textrm{d}}$ \cite{Babenko2014} & 4k & & &  & 54.5 & \textbf{79.3} &  N/A \\ \hline
\textit{Caffe} \cite{Jia13caffe} & 4k & & & & 38.2 & 73 & 3.7 \\ \hline
\textit{Caffe} & 4k & x & x & & 71.2 & 87.1 & 6.4 \\ \hline
\multicolumn{8}{|c|}{Proposed CNN methods} \\ \hline
POI-CNN-N & 4k & & & & 65.7  & 77 & 10.3 \\ \hline
POI-CNN-A & 4k & & & & 64.6  & 76.5& 9.8 \\ \hline
POI-CNN-W & 4k & & & & 67.1  & 76.3 & 9.8 \\ \hline

POI-CNN-N$_{512}$ & 512 & & & & 68.6 & 78 & 11.9 \\ \hline
POI-CNN-A$_{512}$ & 512 & & & & 66.4 & 77.9 & 10.9 \\ \hline
POI-CNN-W$_{512}$ & 512 & & & & \textbf{69.3} & 77.7 & 10.8 \\ \hline


POI-CNN-N$_{512}$ & 3-16k  & x & x &  & \textbf{76.9} & \textbf{87.5} & \textbf{12.5} \\ \hline
POI-CNN-A$_{512}$ & 3-16k & x & x &  & 76.7 & 86.9 & 11.5 \\ \hline
POI-CNN-W$_{512}$ & 3-16k & x & x &  & 76.7 & 86.6 & 11.8 \\ \hline

\end{tabular}}
\end{table}
The results obtained with the existing high-performance models and with the proposed approach are summarized in Table~\ref{tab:results}.
The main finding is that the adapted CNN features introduced here perform better than generic \verb|Caffe| features for all three datasets, a result which indicates that domain adaptation is effective. 
Differences are particularly high for \textit{Oxford5k} and \textit{Div150Cred} datasets, with improvement of over 100\% and respectively 200\% over the generic \verb|Caffe| features.
Equally important, we obtain state-of-the art performance among CNN on both Oxford5k and Holidays. 
Compared to the best non-CNN methods, adapted features still lag behind fine-tuned high-dimensional features~\cite{Arandjelovic12}, \cite{Tolias:2013} for \textit{Oxford5k} and has comparable performance on \textit{Holidays}.
It is noteworthy that the results reported in~\cite{Arandjelovic12}, \cite{Tolias:2013} are obtained with features which are less scalable than ours. 
The adapted CNN features compare favorably with the best configuration of the recently proposed triangulation embedding~\cite{Jegou:2014}, for both \textit{Oxford5k} and \textit{Holidays}. This comparison is interesting because the two types of features are comparable from a scalability point of view. 
BoVW with a very large vocabulary is another strong baseline and a comparison to it shows that the best configurations of adapted CNN features outperform it on all three test datasets. 
The gain is particularly important for Div150Cred, where the mAP score goes from 2.71\% to 12.5\%. 
This result is important from an application perspective since the Div150Cred dataset is the closest proxy to real Web tourism corpora.

Surprisingly, the results obtained with POI-CNN-N, \textit{i.e.}\ direct use of Flickr images for POIs, are comparable, when not better, to those obtained with automatic image reranking and weakly supervised reranking (POI-CNN-A and POI-CNN-W, respectively). 
A qualitative evaluation of reranking results indicates that the two proposed methods are effective in cleaning noise but also decrease the overall diversity of the images.
An empirical explanation of the comparable results of the three methods is that, up to a certain level, training set noise is compensated by diversity.
In other words, a diversified negative set is at least as important as a clean positive set. 
If confirmed beyond the tourism domain, this finding has important implications since CNN training could be done without cumbersome annotation of large volumes of data.

\subsection{Feature compression effect}
We apply a standard PCA-based compression of the adapted CNN features in order to improve their scalability and present the obtained performance in Table~\ref{tab:pca_effect} for \textit{Oxford5k} and \textit{Holidays} datasets. 
Overall, the best results are obtained when retaining the first 256 components of the PCA vectors and this shows that compression is actually beneficial when applied to adapted CNN features.
Below 256 dimensions, performances start to drop but interesting results are obtained even when retaining as few as 16 dimensions. 
For instance, on Oxford5k, a mAP of 53.9\% is similar to~\cite{Babenko2014} (54.5\%) for POI-adapted CNNs with 4k dimensions. 
The behavior of the features proposed here with respect to compression shows that they provide a good balance between scalability and efficiency and could thus be exploited to mine Web-scale datasets. 

\begin{table}[tb!]
\centering
\caption{Effect of PCA-based dimensionality reduction. Results are presented for adapted CNN features obtained without reranking (POI-CNN-N).}
\label{tab:pca_effect}
\resizebox{0.48\textwidth}{!}{
\begin{tabular}{|l|c|c|c|c|c|c|c|c|} 
\cline{2-9}
\multicolumn{1}{c|}{} & \multicolumn{8}{|c|}{\textbf{Dimension}} \\ \cline{2-9}
\multicolumn{1}{c|}{} & \textbf{4096} & \textbf{1024} & \textbf{512} & \textbf{256} & \textbf{128} & \textbf{64} & \textbf{32} & \textbf{16} \\ \hline
Ox5k & 65.7 & 67.8 & 68.6 & 69.2 & 68.6 & 66.2 & 62.2 & 53.9 \\ \hline
Holi & 77 	& 76.8 & 78 & 78.5 & 77.8 & 75.5 & 71.3 & 63.9 \\ \hline
\end{tabular}}
\end{table}

\subsection{Feature tuning effect} 
Confirming results presented in~\cite{DBLP:journals/corr/RazavianASC14}, spatial search and data augmentation bring further improvement compared to the direct use of full image features. 
Relative mAP improvements of over 10\% are obtained for \textit{Oxford5k} and \textit{Holidays} and between 5\% and 10\% for \textit{Div150Cred}. 
The lower improvement for the latter dataset is explained by the fact that the initial image size is smaller (500 pixels on the larger dimension) and only two levels were used for spatial search. 
The best previous retrieval results are obtained with generic features, combined with spatial search and feature augmentation \cite{DBLP:journals/corr/RazavianASC14} using \verb|OverFeat|. We repeated these experiments with \verb|Caffe| and results reach $71.2\%$, $87.1\%$ and $6.4\%$ on \textit{Oxford5k}, \textit{Holidays} and \textit{Div150Cred}, respectively. 
While generic feature tuning effectiveness gives good results on the first two datasets, it clearly fails to match the performance of domain adaptation on the more complex \textit{Div150Cred}.

In addition to spatial search and data augmentation presented in Table~\ref{tab:results}, query expansion has a beneficial effect for \textit{Oxford5k}. We adopt an average query expansion method using the top 20 results from an initial list as positive samples for building an improved query. We therefore reach a mAP of 81.5\%, 80.9\% and 82.7\% for unsupervised, automatic reranking and weakly supervised reranking models respectively.
However, this form of query expansion didn't have a positive effect on \textit{Holidays} and \textit{Div150Cred}.

\section{Conclusions} 
\label{sec:conclusions}
We proposed a CNN adaptation pipeline which learns features from an domain-adapted but noisy Web dataset and successfully transfers them for retrieval in the target domain. 
One important contribution is to show that learning is effective even in presence of noisy images. 
If confirmed in other domains, this finding has important implications since it indicates that manually created resources, such as ImageNet, could be successfully replaced with Web datasets for image retrieval.
Naturally, the quality of the obtained representations depends on an appropriate choice of domain concepts and on the quality of the underlying image source. 

A second contribution is to show that domain adaptation of CNN features is efficient. Compared to~\cite{Babenko2014}, we obtain better results while requiring no manual supervision and also by a more careful setting of the learning process. In addition, our results regarding the usefulness of feature tuning confirm those of~\cite{DBLP:journals/corr/RazavianASC14}.

The results reported here are very encouraging and we intend to pursue work in different inter-related directions.
Experiments were performed with datasets from the tourist domain and one important future work axis is to confirm them on other domains. 
Another is to test more complex CNN architectures~\cite{DBLP:journals/corr/SimonyanZ14a} and to feed the network with larger datasets (\textit{i.e.}\ more concepts and more images per concepts).
Finally, building on~\cite{gong:2014}, we want to explore the alteration of CNN architectures in order to better account for spatial context.

\bibliographystyle{IEEEbib}
\bibliography{cnn_adaptation}

\end{document}